\documentclass[twoside,leqno,twocolumn]{article}

\usepackage[letterpaper]{geometry}

\usepackage{ltexpprt}

\usepackage[utf8]{inputenc} 
\usepackage[T1]{fontenc}    
\usepackage{hyperref}       
\usepackage{url}            
\usepackage{booktabs}       
\usepackage{amsfonts}       
\usepackage{nicefrac}       
\usepackage{microtype}      

\usepackage{graphicx, caption, subcaption}

\usepackage{algorithm}
\usepackage{algorithmic}
\usepackage{multirow}
\usepackage{amssymb}
\usepackage{amsmath}
\usepackage{wasysym}
\usepackage{enumitem}
\usepackage[T1]{fontenc}
\usepackage{mathtools, nccmath}
\usepackage{bbm}
\usepackage{pifont}
\usepackage{amssymb}
\DeclareMathOperator*{\argmax}{arg\,max}

\usepackage{soul}
\usepackage{xcolor}

\makeatletter
\def\thickhline{%
	\noalign{\ifnum0=`}\fi\hrule \@height \thickarrayrulewidth \futurelet
	\reserved@a\@xthickhline}
\def\@xthickhline{\ifx\reserved@a\thickhline
	\vskip\doublerulesep
	\vskip-\thickarrayrulewidth
	\fi
	\ifnum0=`{\fi}}
\makeatother

\newlength{\thickarrayrulewidth}
\begin{document}

\title{\Large Controlled Molecule Generator 
	\\ for Optimizing Multiple Chemical Properties
}
\author{Bonggun Shin\thanks{Deargen Inc., South Korea. Email: \href{mailto:bonggun.shin@deargen.me}{bonggun.shin@deargen.me} }
	\and Sungsoo Park\thanks{Deargen Inc., South Korea. Email: \href{mailto:sspark@deargen.me}{sspark@deargen.me}}
	\and JinYeong Bak \thanks{College of Computing at SungKyunKwan University, South Korea. Email: \href{mailto:jy.bak@skku.edu}{jy.bak@skku.edu}}
	\and Joyce C. Ho\thanks{Department of Computer Science at Emory University, Atlanta, GA. Email: \href{mailto:joyce.c.ho@emory.edu}{joyce.c.ho@emory.edu}}}

\date{}

\maketitle


\fancyfoot[R]{\scriptsize{Copyright \textcopyright\ 2021 by SIAM\\
Unauthorized reproduction of this article is prohibited}}





\begin{abstract} \small\baselineskip=9pt 
Generating a novel and optimized molecule with desired chemical properties is an essential part of the drug discovery process. Failure to meet one of the required properties can frequently lead to failure in a clinical test which is costly. In addition, optimizing these multiple properties is a challenging task because the optimization of one property is prone to changing other properties. In this paper, we pose this multi-property optimization problem as a sequence translation process and propose a new optimized molecule generator model based on the Transformer with two constraint networks: property prediction and similarity prediction. We further improve the model by incorporating score predictions from these constraint networks in a modified beam search algorithm. The experiments demonstrate that our proposed model outperforms state-of-the-art models by a significant margin for optimizing multiple properties simultaneously.
\end{abstract}

\section{Introduction}
\label{sec:introduction}

Drug discovery is an expensive process. According to Dimasi et al.~\cite{dimasi2016innovation}, the estimated average cost to develop a new medicine and gain FDA approval is \$1.4 billion. Among this amount, 40\% of it is spent on the candidate compound generation step. In this step, around 5,000 to 10,000 molecules are generated as candidates but 99.9\% of them will be eventually discarded and only 0.1\% of them will be approved to the market.
This inefficient nature of the candidate generation step serves as motivation to design an automated molecule search method.
However, finding target molecules with the desired chemical properties is challenging because of two reasons.
First, an efficient search is not possible because the search space is discrete to the input~\cite{kirkpatrick2004chemical}.
Second, the search space is too large that it reaches up to $10^{60}$~\cite{polishchuk2013estimation}.
As such, this task is currently being tackled by pharmaceutical experts and takes years to design.
Therefore, this paper aims to accelerate the drug discovery process by proposing a deep-learning (DL) model that accomplishes this task effectively and quickly.

Recently, many methods of molecular design have been proposed~\cite{bjerrum2017molecular,segler2018generating,ertl2017silico,gomez2018automatic,dai2018syntax,kusner2017grammar,olivecrona2017molecular,guimaraes2017objective,sanchez2017optimizing,yang2017chemts}. 
Among them, Matched Molecular Pair Analysis (MMPA)~\cite{griffen2011matched} and Variational Junction Tree Encoder-Decoder (VJTNN)~\cite{jin2019learning} formulated molecular property optimization as a problem of molecular paraphrase.
Just as a Natural Language Process (NLP) model produces paraphrased sentences, when a molecule comes in as an input to these models, another molecule with improved properties is generated by paraphrase.
Although MMPA was the first to try this approach, it is not effective unless many rules are given to the model~\cite{jin2019learning}.
To mitigate this problem, Jin et al.~\cite{jin2019learning} proposed VJTNN, an end-to-end molecule optimization model without the need for rules.
By efficiently encoding and decoding a molecule with graphs and trees, it is the current state-of-the-art (SOTA) model for optimizing a single property (hereby referred to as a single-objective optimization task).
However, it cannot optimize multiple properties at the same time (a multi-objective optimization task) because the model inherently optimizes only one property. 
As noted by Shanmugasundaram et al.~\cite{shanmugasundaram2016monitoring} and Vogt et al. ~\cite{vogt2018computational}, the actual drug discovery process frequently requires balancing of multiple properties.

With these motivations, we propose a new DL-based end-to-end model that can \emph{optimize multiple properties in one model}. By extending the preceding problem formulation, we consider the molecular optimization task as a sequence-based controlled paraphrase (or translation) problem.  
The proposed model, controlled molecule generator (CMG), learns how to translate the input molecules given as sequences into new molecules as sequences that best reflect the newly desired molecule properties.
Our model extends the Transformer model~\cite{vaswani2017attention} that showed its effectiveness in machine translation.
CMG encodes raw sequences through a deep network and decodes a new molecule sequence by referencing that encoding and the desired properties. Since we represent the desired properties as a vector, this model inherently can consider multiple objectives simultaneously. Moreover, we present a novel loss function using pre-trained constraint networks to minimize generating invalid molecules. Lastly, we propose a novel beam search algorithm that incorporates these constraint networks into the beam search algorithm~\cite{medress1977speech}.

We evaluate the proposed model using two tasks (single-objective optimization and multi-objective optimization) and two analysis studies (ablation study case study)\footnote{Code and data are available at \url{https://github.com/deargen/cmg}}.
We compare our model with six existing approaches including the current SOTA, VJTNN.
CMG outperforms all baseline models in both benchmarks.
In addition, our model is trained once and evaluated for all tasks, which shows practicality and generality.
The ablation study not only shows the effectiveness of each sub-part, but demonstrates the superiority of the proposed model itself without the sub-parts.
Lastly, the case study demonstrates the practicality of our method through the target affinity optimization experiment using an actual experimental drug molecule.

The contributions of this paper are 
(1) a new formulation of the multi-objective molecule optimization task as a sequence-based controlled molecule translation problem,
(2) a new self-attention based molecule translation model that can reflect the multiple desired properties through constraint networks,
(3) new loss functions to incorporate the pre-trained constraint networks,
and (4) a novel beam search algorithm using the pre-trained constraint networks.

\section{Related Work}
\label{sec:related.work}
\textbf{Molecule property optimization}:
Molecule property optimization models can be divided into two types depending on the data representation: sequence representations and graph representations. One of the earlier approaches using sequence representations utilizes encoding rules~\cite{weininger1988smiles}, while the recent ones~\cite{gomez2018automatic,segler2018generating,kusner2017grammar} are based on DL methods that learn to reconstruct the input molecule sequence.
This is related to our work in terms of the input representation, but they offer subpar performance when compared to the SOTA models.
Another group of research uses graph representations conveying structural information~\cite{dai2016discriminative,jin2018junction,samanta2018designing,li2018multi,dalke2018mmpdb}.
Among them, VJTNN~\cite{jin2019learning} and MMPA~\cite{griffen2011matched,dossetter2013matched,dalke2018mmpdb} are closely related to our work
because they formulate the molecule property optimization task as a molecule translation problem. 
From the model perspective, MMPA is a rule-based model and VJTNN is a supervised DL model. 
Although our approach is also based on a DL method, there is a big difference in practical use cases.
A single VJTNN model is capable of optimizing a single property, while the proposed model can optimize multiple properties by using the controlled decoder.
With these differences, we formulate the molecule property optimization task as a ``controlled'' molecule ``sequence'' translation problem. 
Other molecule generation methods include
Junction Tree Variatinal Auto Encoder (JT-VAE)~\cite{jin2017predicting},
Variational Sequence-to-Sequence (VSeq2Seq)~\cite{gomez2018automatic,bahdanau2014neural},
Graph Convolutional Policy Network (GCPN)~\cite{you2018graph},
and Molecule Deep Q-Networks (MolDQN)~\cite{zhou2019optimization}.

\noindent\textbf{Natural Language Generation Model}:
Our model is inspired by the recent success in molecule representation using the self-attention technique~\cite{pmlr-v106-shin19a}. By adopting the BERT~\cite{devlin2018bert} architecture to represent molecule sequences, their model becomes the SOTA in the drug-target interaction task. 
In terms of the model architecture, our work is related to Transformer~\cite{vaswani2017attention} because we extend it to be applicable to the molecule optimization task.
There is a controlled text generation model~\cite{hu2017toward} in NLP domain. It is related to ours because they feed the desired text property as one of the inputs. 
	However, all of these methods are designed for NLP tasks, therefore, they cannot be directly applied to molecule optimization tasks for two reasons. Firstly, the similarity constraint of the molecule optimization is an important feature, however, a typical NLP model can't reflect this. Secondly, NLP models take categorical properties while ours is designed for numerical ones, which is more realistic in a molecule optimization.

\noindent\textbf{Transfer learning}:
DL-based transfer learning by pre-training has been applied to many fields such as computer vision~\cite{rothe2015dex,ghifary2016deep}, NLP~\cite{howard2018universal}, speech recognition~\cite{jaitly2012application,lu2013speech}, and health-care applications~\cite{shin2017classification}.
They are related to ours because we also pre-train the constrained networks and transfer the weights to the main model.

\section{Controlled Molecule Generator}
\label{sec:methods}




\begin{figure*}[ht]
	\centering
	\captionsetup{type=figure} 
	\centering
	\subcaptionbox{Training. Reds represent the losses.\label{fig:model.overview.a}}
	{\includegraphics[width=0.43\textwidth]{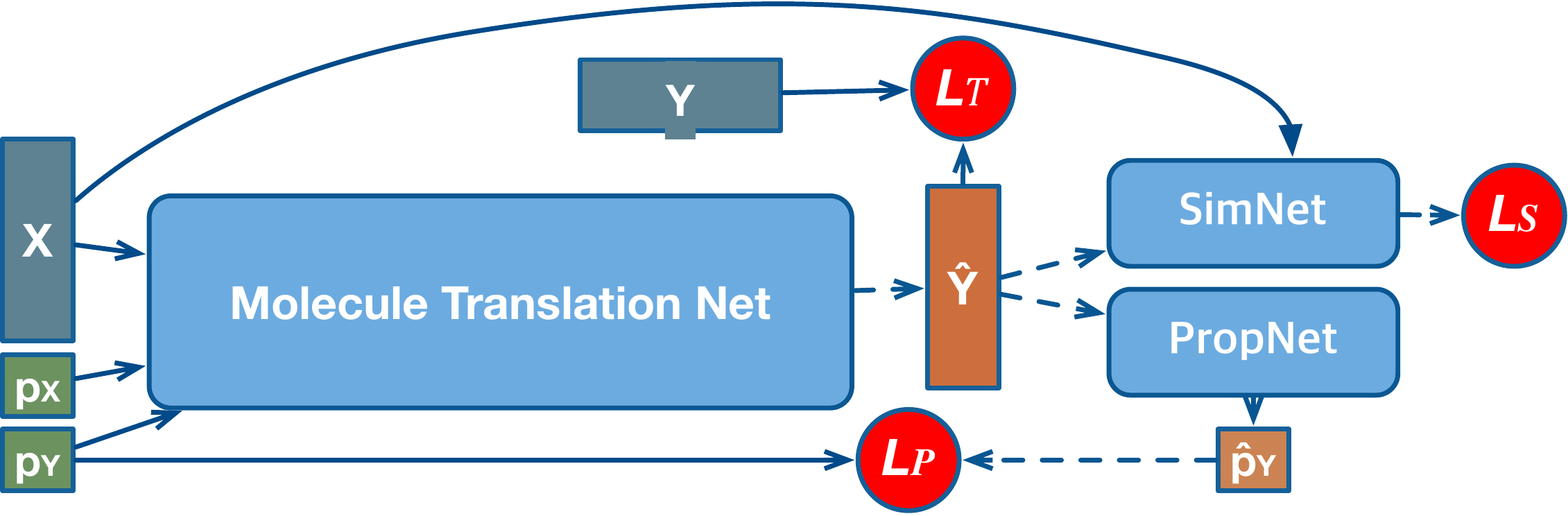}}\quad
	\subcaptionbox{Prediction using the constraint nets.\label{fig:model.overview.b}}
	{\includegraphics[width=0.53\textwidth]{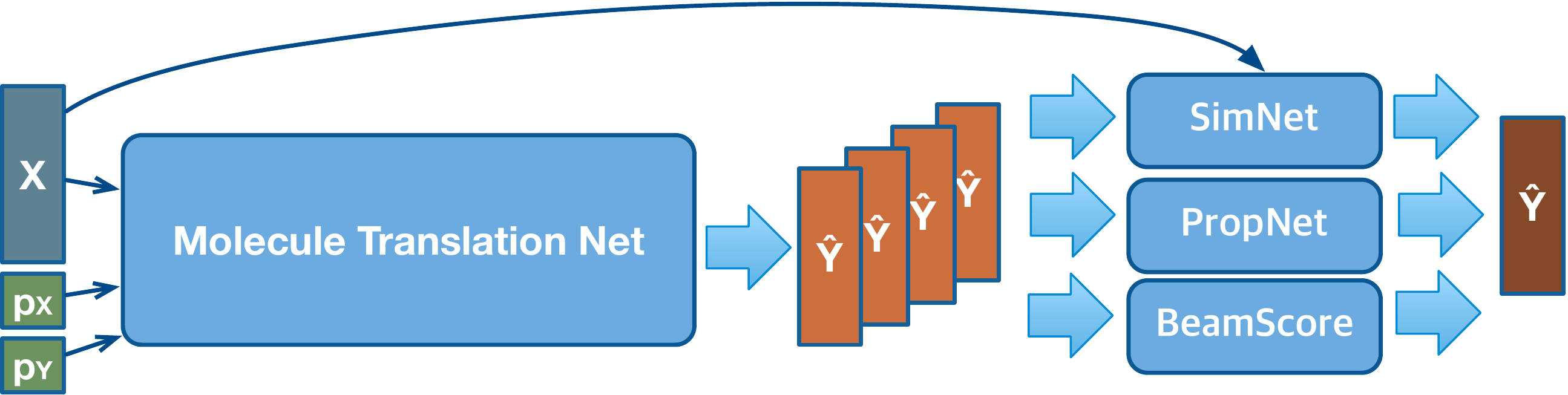}}
	\caption{The proposed controlled molecule generator model at training and prediction.}
	\label{fig:model.overview}
\end{figure*}


\subsection{Problem Definition}
Given an input molecule $X$, its associated molecule property vector $p_X$, and the desired property vector $p_Y$, the goal is to generate a new molecule $Y$ with the property $p_Y$ with the similarity of $(X,Y) \ge \delta$. Note that $\delta$ is a similarity threshold and the similarity measure is Tanimoto molecular similarity over Morgan fingerprints~\cite{rogers2010extended}.
Formally, for two Morgan fingerprints, $F_X$ and $F_Y$, where both of them are binary vectors, the Tanimoto molecular similarity is  $sim(F_X,F_Y) = \frac{|F_X \cap F_Y|}{|F_X \cup F_Y|}$.


\vspace{-0.5em}
\subsection{Model Overview}
\vspace{-0.5em}
Our model extends the Transformer~\cite{vaswani2017attention} to a molecular sequence by incorporating molecule properties and additional regularization networks. 
Inspired by the previous success in applying the self-attention mechanism to represent a molecule sequence~\cite{pmlr-v106-shin19a}, we treat 
each molecule just like a sequence. 
However, this NLP technique cannot be directly applied, because the structure of the molecular sequence differs from natural languages, where the hierarchy is a letter-word-sentence.
Not only that, there is no training data available that is collected for the molecule translation task, while there are ample datasets in the NLP domain.
To fill these gaps, we propose the controlled molecule generation model (Figure~\ref{fig:model.overview}) and present how we gather the training data for this network (Section~\ref{ssec:datasets}).
We optimize the proposed model using three loss functions as briefly shown in Figure~\ref{fig:model.overview.a}.
In addition, we propose 
two constraint networks (Section~\ref{ssec:constraint.networks}, Figure~\ref{fig:model.constnets}), the 
property prediction network and the similarity prediction network to train the model more accurately.
Lastly, we also present how we modify the beam search algorithm~\cite{medress1977speech} to best exploit the existing auxiliary networks,
as briefly shown in Section~\ref{ssec:beamsearch} and Figure~\ref{fig:model.overview.b}.

\vspace{-0.5em}
\subsection{Molecule Translation Network}
\label{ssec:mol.translation}
\begin{figure}
	\centering
	\includegraphics[width=0.48\columnwidth]{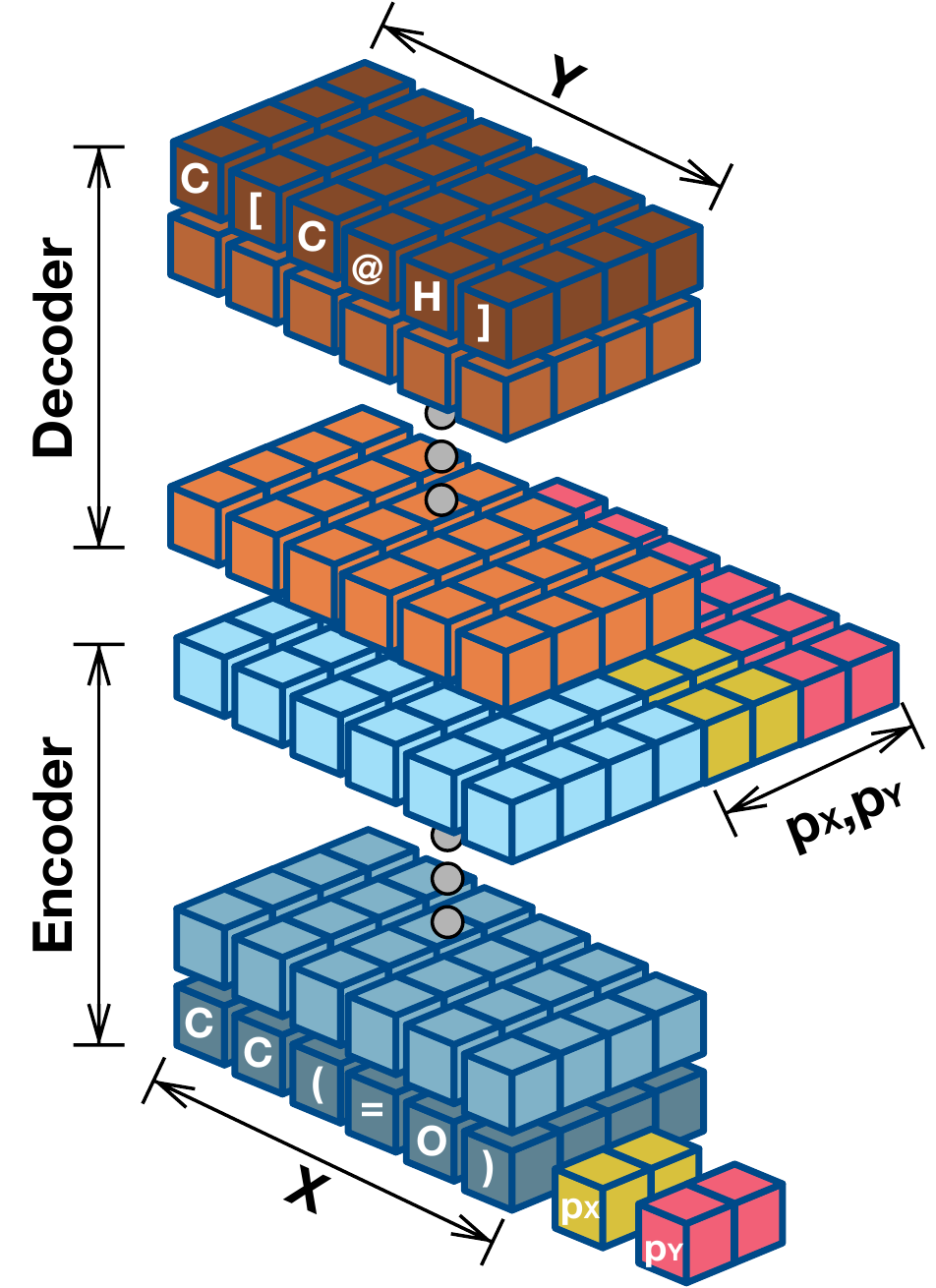}
	\caption{The molecule translation network.}  
	\label{fig:model.trans}
\end{figure}

\vspace{-0.5em}
We apply two modifications to the Transformer model~\cite{vaswani2017attention}.
First, unlike Transformer which uses word embeddings, we use character embeddings because 
the molecule sequence is comprised of characters representing atoms or structure indicators. 
To mark the beginning and the end of a sequence, we add ``[BEGIN]'' token at the first position of the sequence
and ``[END]'' token at the last. 
Another modification is that we add chemical property awareness to the hidden layer of the Transformer model.
We enrich token vectors of the last encoder by concatenating property vectors to each of the token vectors as shown in Figure~\ref{fig:model.trans}.
Formally, let $z_i$ be the token vectors in the last encoder.
Then, the new encoding vector becomes $z'_i = (z_i , p_X , p_Y)  \in \mathbb{R}^{d+2k}$, where $k$ represents the number of properties.
Although it might be seen as a simple method, this empirically shows the best result among other types of configurations,
such as property embeddings, disentangled encodings (property and non-property encodings), 
and concatenating property differential information instead of providing two raw vectors.
The cost function of this network is the cross entropy between the target ($y_i$) and predicted molecule ($\hat{y_i}$). 
Therefore, it is formally defined as $\mathcal{L}_T(\boldsymbol{\theta}_T;X, p_X, p_Y) = -\frac{1}{N}\frac{1}{M}\sum_{n \in N}\sum_{j \in M}\sum_{v \in V} y_{v,j,n} \cdot \log(\hat{y}_{v,j,n})$, 
where $\boldsymbol{\theta}_T$ denotes all parameters of the Transformer and $N$, $M$, $V$ represent the number of training samples, the length of a sequence, and the size of the vocabulary, respectively.

\subsection{Constraint Networks}
\label{ssec:constraint.networks}

\begin{figure}[thb]
	\centering
	\includegraphics[width=0.75\columnwidth]{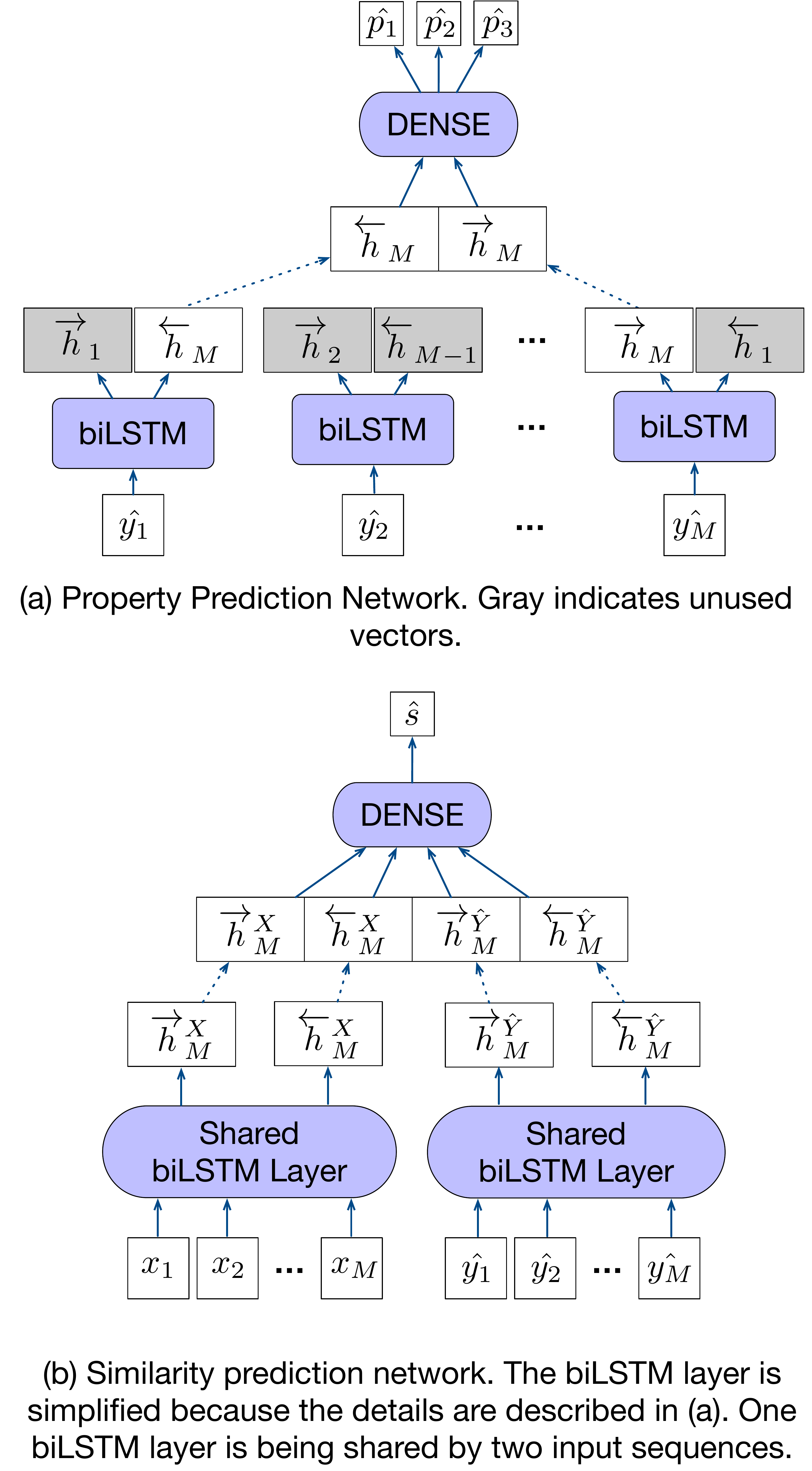}
	\caption{Two Constraint Networks.}
	\label{fig:model.constnets} 
\end{figure}

We hypothesize that the cost function of the Transformer network ($\mathcal{L}_T$) is not enough to teach the generating model, because the error signals from this loss function can hardly capture the valuable information, such as if a predicted sequence pertains to the desired property or if it satisfies the similarity constraint.
With this motivation, we add two constraint networks as follows.


\subsubsection{Property Prediction Network}
\label{sssec:prop.pred.net}
\vspace{-0.5em}
The property prediction network (PropNet) takes the predicted molecule sequence ($\hat{y}_i$) as an input (on the top of Figure~\ref{fig:model.constnets}). 
The left-to-right LSTM~\cite{hochreiter1997long} layer and the right-to-left one encode input vectors ($\hat{y}_i$) into hidden vectors, $\overrightarrow{h}_i \in \mathbb{R}^d$ and $\overleftarrow{h}_j \in \mathbb{R}^d$, respectively.
Since the last vectors for each direction summarize the sequence,
they are concatenated ($h_{prop}=(\overrightarrow{h}_M,\overleftarrow{h}_M) \in \mathbb{R}^{2d}$) and fed into a dense network with two hidden layers.
With the predicted property, $\hat{p}_Y$ and the desired property ($p_Y$) from the input, we can create a loss function, 
which will enrich error signals by adding property awareness in predicting a molecule.
This is formally written as, $\mathcal{L}_P(\boldsymbol{\theta}_T;X, p_X, p_Y) = \frac{1}{N} \sum_{n\in N} |p_{Y_n}-\hat{p}_{Y_n}|^2$.
We pre-train PropNet using molecules in the training set, and the properties are calculated using a third-party library.
Once pre-trained, the parameters are transferred to the CMG network and frozen when training the CMG.

\subsubsection{Similarity Prediction Network}
\label{sssec:sim.pred.net}
The input of the similarity prediction network (SimNet) is composed of the predicted molecule sequence ($\hat{y}_i$) and the input molecule sequence ($x_i$) (on the bottom of Figure~\ref{fig:model.constnets}).
We posit that adding estimated similarity error signals to the loss function could be useful for satisfying the similarity requirement because 
this is relatively direct information in the generation modeling.
We employ one layer of BiLSTM for SimNet, which is shared by two different inputs.
Two inputs ($\hat{y}_i$ and $x_i$) is passed to the BiLSTM layer to produce each corresponding feature vector, $h^{\hat{Y}}_M \in \mathbb{R}^{2d}$ and $h^{X}_M \in \mathbb{R}^{2d}$ ($M$ indicates the last token index).
We concatenate these two feature vectors as $(h^{\hat{Y}}_M,h^{X}_M) \in \mathbb{R}^{4d}$, so that the next two dense networks can capture the similarity between the two.
After applying two-layered dense network, we get the binary prediction ($\hat{s}_n$) whether the two input molecules are similar or not according to the threshold $\delta$.
With this prediction ($\hat{s}_n$) and the label ($s_n$), 
we can create the last loss function, formally written as 
$\mathcal{L}_S(\boldsymbol{\theta}_T;X, p_X, p_Y) = \frac{1}{N} \sum_{n\in N} s_n \log{\hat{s}_n} +(1-s_n) \log{(1-\hat{s}_n)}$.
We transfer the pre-trained SimNet weights into the CMG model and freeze the SimNet weights when training the CMG network.


\subsubsection{CMG Loss Function}
\label{sssec:loss.function}

By combining all cost functions ($\mathcal{L}_{T}, \mathcal{L}_{P}, \mathcal{L}_{S}$), we can obtain the CMG loss function as 
$\displaystyle \mathcal{L}_{CMG} = \mathcal{L}_{T} + \lambda_p \mathcal{L}_{P} + \lambda_s \mathcal{L}_{S},$
, where $\lambda_p$ and $\lambda_s$ are weight parameters.

\subsection{Modified Beam Search with Constraint Networks}
\label{ssec:beamsearch}

\begin{algorithm}[tb]
	\begin{algorithmic}[1]
		\STATE {\bfseries Input:} Candidate molecules: $C_1, C_2, \cdots, C_b$, \\
		\;\;\;\;\;\;\;\;\;\; Corresponding beam scores: $s_1, s_2, \cdots, s_b$ \\
		\;\;\;\;\;\;\;\;\;\; Input molecule: $X$ \\
		\;\;\;\;\;\;\;\;\;\; Desired property vector: $p_Y$
		\FOR{$i=1$ {\bfseries to} $b$}
		\STATE $\hat{p}_i\leftarrow$ PropNet($C_i$)
		\STATE $p_{d} \leftarrow |p_Y-\hat{p}_i|$
		\STATE $s_{pn}\leftarrow$ reduce\_mean($1-p_d$)
		\STATE $s_{sn}\leftarrow$ SimNet($X, C_i$)
		\STATE $s_i \leftarrow s_i+ (s_{pn}+s_{sn})$
		\ENDFOR
		\STATE best\_index $\leftarrow \argmax s_i$
		\STATE {\bfseries Output:} $C_{\text{best\_index}}$
	\end{algorithmic}
	\caption{Modified Beam Search}
	\label{alg:mbs}
\end{algorithm}

When generating a sequence from CMG at testing, there is no gold output sequence that it can reference.
Therefore we need to sequentially generate tokens until we encounter the "[END]" token, like other sequence-based algorithms.
At this process, a typical way is the beam search, where the model maintains top $b$ number of best candidate sequences
when predicting each token.
When all candidate sequences are complete and ready, the model outputs the best candidate in terms of a beam score, a cumulative log-likelihood score for a corresponding candidate.
However, the standard beam search does not account for the multi-objective nature of our task.
For example, there is a possibility that low ranked molecules could be closer to the desired properties than the top molecule selected by the beam search.
Therefore, we propose a modified beam search algorithm (Algorithm~\ref{alg:mbs}) using our constraint networks.
For the property evaluation, we first get the predicted property of each candidate and get the absolute difference from the desired property (Line 3-4 in Algorithm~\ref{alg:mbs}).
Since this difference is desired to be small, we calculate the property evaluation score ($s_{pn}$) by subtracting them from one (Line 5 in Algorithm~\ref{alg:mbs}).
The property could have multiple values, therefore, we take an average of all elements of this difference vector.
For the similarity evaluation, we get the predicted similarity between the input $X$ and each candidate $C_i$ (Line 6 in Algorithm~\ref{alg:mbs}).
Since we expect a candidate should be similar (label 1) to the input, we regard the predicted similarity as the raw score from SimNet.
By adding these two predicted scores to the original beam scores, we obtain the modified beam scores (Line 7 in Algorithm~\ref{alg:mbs}).
With this new score, we can select the best candidate (Line 9-10 in Algorithm~\ref{alg:mbs}).

\subsection{Diversifying the Output}
\label{ssec:gen.diff.mol}
Unlike other variational models (VSeq2Seq and VJTNN), the proposed one encodes a fixed vector
that is able to generate a single output for one input.
In order to diversify the output for a fixed input, 
we re-parameterize the desired vector, $(p_1, p_2, p_3)$, as random variable by adding a Gaussian noise with a user-specified variance, 
$\Tilde{p}_k \sim N(p_k, \sigma_k)$.
For example, if the desired property vector is $(p_1, p_2, p_3)$, 
we feed $(p_1+\alpha, p_2+\beta, p_3+\gamma)$, where $\alpha, \beta$ and $\gamma$ are samples drawn from $N(0, \sigma_1)$, $N(0, \sigma_2)$, and $N(0, \sigma_3)$.

\section{Experiments}
\label{sec:experiments}
\vspace{-0.5em}
We compare the proposed model with state-of-the-art molecule optimization methods in the following tasks.
\textbf{Single Objective Optimization (SOO)}: This task is to optimize an input molecule to have a better property while preserving a certain level of similarity between the input molecule and the optimized one. Since developing a new drug usually starts with an existing molecule~\cite{bickerton2012quantifying}, this task serves as a good benchmark. 
\textbf{Multi-Objective Optimization (MOO)}: This task reflects a more practical scenario in drug discovery, where modifying an existing drug involves optimizing multiple properties simultaneously, such as similarity, lipophilicity scores, drug likeness scores, and target affinity scores. Since improving one property might often result in sacrificing other properties, this task is harder than a single-objective optimization task.
To present multi-faceted aspects of the proposed model, we additionally perform the following experiments.
\textbf{Ablation Study}: For ablation study, we report the validity of the constraint networks both in training and testing phases.
\textbf{Case Study}: To evaluate the effectiveness of the proposed model, we present the result of an actual drug optimization task with an existing molecule in an experimental phase.
More precise information regarding the reproducibility can be found in the supplementary material.

\subsection{Datasets}
\label{ssec:datasets}
Since CMG is based on sequence translation, we need to appropriately curate the dataset. 
\textbf{Training Set for CMG}:
We use the ZINC dataset~\cite{sterling2015zinc} (249,455 molecules) 
and the DRD2 related molecule dataset (DRD2D)~\cite{olivecrona2017molecular} (25,695 molecules),
which result in 260,939 molecules for our experiments.
This is the same set of molecules on which Jin et al.~\cite{jin2019learning} used to evaluate their model (VJTNN). 
From these 260k molecules, we exclude molecules that appear in the development and the test set of VJTNN, 
resulted in 257,565 molecules.
With these molecules, we construct training datasets 
by selecting molecule pairs $(X,Y)$ with the similarity is greater than or equal to 0.4,
following the same procedure in \cite{jin2019learning, dalke2018mmpdb}.
Jin et al.~\cite{jin2019learning} used the small portion of these pairs by excluding all property-decreased molecule pairs. 
The main difference from their curation processes is that we don't have to exclude many property-decreased molecule pairs because our model can extract useful information even from them.
By doing this, we provide a more ample dataset to a deep model, so that it could be helpful in finding more useful patterns. As a result, the number of pairs in training data is significantly bigger than theirs. 
Among all possible pairs ($257\text{K}\times257\text{K}=67\text{B}$), 
we select 10,827,615 pairs that satisfies similarity condition ($\ge 0.4$).
With the same similarity condition, Jin et al.~\cite{jin2019learning} gathered less than $100$K due to additional constraints of training sets, 
which is only property increased molecule pairs can be used as training data.
As previous works~\cite{jin2019learning,kusner2017grammar} did, 
we pre-calculate the three chemical properties of all molecules ($p_X$ and $p_Y$) in the training set: 
\textbf{Penalized logP (PlogP)~\cite{kusner2017grammar}} is a measure of lipophilicity of a compound, specifically, the octanol/water partition coefficient (logP) penalized by the ring size and synthetic accessibility.
\textbf{Drug likeness (QED)} is the quantitative estimate of drug-likeness proposed by Bickerton et al.~\cite{bickerton2012quantifying}
and \textbf{Dopamine Receptor (DRD2)} is a measure of molecule activity against a biological target, the dopamine type 2 receptor.

\textbf{Training Set for PropNet}:
Among 260,939 molecules, we excluded all molecules in the test sets of the two tasks; single-objective optimization, multi-objective optimization. The number of these remained molecules is 257,565.
We construct the dataset for PropNet by arranging all molecules as inputs and the corresponding three properties as outputs.
We randomly split this into the training and validation sets with a ratio of 8:2. 

\textbf{Training Set for SimNet}:
We use a subset of all 10,827,615 pairs in the CMG training set due to the simpler network configuration of SimNet.
When sub-sampling pairs, we tried to preserve the proportion of the similarity in the CMG dataset to best preserve the original data distribution. 
The reason behind this effort is that preserving the similarity distribution could possibly contribute to the SimNet accuracy although SimNet only uses binary labels.
In addition, we try to preserve the similar/not-similar ratio to be about to the same.
By sampling about 10\% of data, we gathered 997,773 number of pairs and the ratio of the positive samples is 49.45\%.
We randomly split this into the training and validation set with a ratio of 8:2. 


\vspace{-0.5em}
\subsection{Pre-Training of Constraint Networks}
\label{ssec:pretraining}
\vspace{-0.5em}
We pre-train the two constraint networks using the training sets described in Section~\ref{ssec:datasets}.
We choose the pre-trained PropNet that recorded the mean square error of 0.0855 and
the the pre-trained SimNet of 0.9759 accuracy through the best model evaluated on each corresponding development set.
The pre-trained weights of the two networks are transferred to the corresponding part in the CMG model and frozen when training CMG and predicting a new molecule using it.
The details of the configuration is described in the supplement material.

\vspace{-0.5em}
\subsection{Single Objective Optimization}
\label{ssec:soo}
\vspace{-0.5em}
The first task is the single objective optimization task
proposed by Jin et al.~\cite{jin2017predicting}. 
The goal is to generate a new molecule with an improved single property score under the similarity constraint ($\delta=0.4$).
We used the same development and test sets provided by Jin et al.~\cite{jin2017predicting}.
Our model is trained once and evaluated for all tasks (SOO, MOO and the case study).

\textbf{Baselines}: We compare the proposed method with the following baselines;
MMPA, JT-VAE, GCPN, VSeq2Seq, MolDQN, and VJTNN introduced in Section~\ref{sec:related.work}.
Since Jin et al.~\cite{jin2019learning} ran and reported almost all of the baseline methods on the single property optimization task (PlogP improvement task) with the same test sets, we cite their experiment results. For MolDQN, which is published after VJTNN, we referenced the scores from MolDQN paper~\cite{zhou2019optimization}.

\textbf{Metrics}:
Since the task is to generate a molecule with an improved PlogP value, we measure an average of raw increments and its standard deviation among valid molecules with the similarity constraint met. Following the VJTNN procedure~\cite{jin2019learning}, one best molecule is selected among 20 generated molecules. 
We also measure the diversity defined by Jin et al.~\cite{jin2019learning}.
Although this diversity measure has been used by previous researches, it is limited in that it encourages the outputs to have low similarity around the threshold. 
However, this could be beneficial in a practical situation where the model needs to generate various molecules around the similarity threshold.


\begin{table}[]
	\centering
	\resizebox{\columnwidth}{!}{%
		\begin{tabular}{lcccc}
			\thickhline
			& \multicolumn{2}{c}{Single Obj.Opt.} & \multicolumn{2}{c}{Multi Obj.Opt.}      \\ \cmidrule(lr){2-3} \cmidrule(lr){4-5}
			Method    & Improvement                   & Diversity          & All Samples &  Sub Samples \\ \hline
			MMPA     & 3.29 $\pm$ 1.12               & 0.496              & -                          & -                         \\
			JT-VAE   & 1.03 $\pm$ 1.39               & -                  & -                          & -                         \\
			GCPN     & 2.49 $\pm$ 1.30               & -                  & -                          & -                         \\
			VSeq2Seq & 3.37 $\pm$ 1.75               & 0.471              & -                          & -                         \\
			MolDQN   & 3.37 $\pm$ 1.62               & -                  & -                          & 0.00\%                    \\
			VJTNN    & 3.55 $\pm$ 1.67               & 0.480              & 3.56\%                     & 4.00\%                    \\ \hline
			Proposed & \textbf{3.92 $\pm$ 1.88}      & \textbf{0.545}     & \textbf{6.98\%}            & \textbf{6.00\%}          \\ \thickhline
	\end{tabular}}
	\caption{Single and multi objective optimization performance comparison on the penalized logP task. For the single one,
		MolDQN results are from~\cite{zhou2019optimization}, and the scores of other baselines are from~\cite{jin2019learning}.
		The reported scores of the multi objective optimization task is a success rate.
	}
	\label{tbl:single.opt}
	\vspace{-1.5em}
\end{table}

\textbf{Result}:
After we train the model 
we generate new molecules by feeding input molecules and desired chemical properties to the trained model.
As discussed in Section~\ref{ssec:gen.diff.mol}, we add offsets to desired properties so that the output can be diversified.
Since the number of generated samples for each input is set to 20, we use the desired property vector of $\{X_{P\log{P}}, 0.0, 0.0\}$
with a total of 20 combinations of ($\alpha$,$\beta$,$\gamma$)  that are sampled from the user-defined distributions.
We select the best model using the development set, and the test set performance of that model is reported in the left part of Table~\ref{tbl:single.opt}.
In the PlogP optimization task, the proposed model outperforms all baselines including the current SOTA, VJTNN,
in terms of both the average improvement and the diversity by a large margin.
Considering the two recently proposed methods (MolDQN and VJTNN) are competing in 0.18 difference, the proposed one surpasses the current SOTA by 0.37 improvement.
The same trend can be found in the diversity comparison.
For QED and DRD2 cases, however, CMG underperforms the others (the scores are in the supplemental material).
The primary reason is that the CMG model is trained once for the MOO task. This model is then re-used and evaluated on the SOO tasks.
More specifically, the proportions of improved QED and DRD2 pairs in the training set are just 5.9\% and 0.08\%, respectively.
Therefore, when optimizing solely for QED or DRD2, CMG could not fully extract the useful information from the training set.
Since our model is trained once for all tasks (SOO, MOO, and the case study), this small portion of information can negatively impact certain single property optimizations, such as QED and DRD2.
However, considering SOO is less practical in drug discovery, the focus should be on the MOO results.

\subsection{Multi Objective Optimization}
\label{ssec:moo}
We set up a new benchmark, multi-objective optimization (MOO) because the actual drug discovery process frequently requires balancing of multiple compound properties \cite{vogt2018computational, shanmugasundaram2016monitoring}.
In this task, we jointly optimize three chemical properties for a given molecule.
We set up the success criteria of the generated molecules in the MOO task as follows:
(1) sim$(X,Y)\ge 0.4$, 
(2) PlogP improvement is at least 1.0,
(3) QED value is at least 0.9
and (4) DRD2 value is over 0.5. 
	We created the above four conditions by combining the existing single optimization benchmarks from VJTNN~\cite{jin2019learning} as one simultaneous condition\footnote{For the PlogP improvement, we set a hard number of 1.0 instead of measuring the magnitude of improvements to transform the criteria into a binary condition.}. 

To create the development set of this task, we merge all three different development sets provided by VJTNN, consisting of 1,038 molecules.
Among those molecules, we exclude any molecules that already satisfy the above criteria.
Then, the final development set contains 985 molecules.
We perform the same procedure for the test set, which reduces the number of molecules to 2,365.


\textbf{Baselines}:
For this task, we include the top two baselines (MolDQN and VJTNN) from the SOO task.
While MolDQN can perform the MOO task by simply modifying the reward function, VJTNN can't perform as it is because it is designed for a single property optimization.
Here are how we prepare those baselines for the MOO task.
\begin{itemize}[leftmargin=.1in]
	\item \textbf{MolDQN}: The reward function of MolDQN for this task is defined as 
	$r=\frac{1}{8}\mathbbm{1}(sim(X,Y)\ge0.4)+ \frac{1}{8}\mathbbm{1}(P\log{P}(Y)-P\log{P}(X)\ge1.0)
	+ \frac{1}{8}\mathbbm{1}(QED \ge 0.9) + \frac{1}{8}\mathbbm{1}(DRD2>0.5) 
	+ \frac{1}{8}\text{sim}(X,Y) + \frac{1}{8}P\log{P}(Y)
	+ \frac{1}{8}\text{QED}(Y) + \frac{1}{8}\text{DRD2}(Y)
	$.
	The first four terms represent the exact goal of the task, and the last four terms provide continuous information about the goals.
		Unlike VJTNN and CMG that require mere evaluation of the trained models, MolDQN should be re-trained from the beginning for each test sample, which requires significant time. Therefore, evaluating MolDQN for all 2,365 samples requires 2,365 times of training that is estimated as more than three months with a 96-CPUs server.
	
	Therefore, we sub-sample the test set ($n=50$) while preserving the original distribution\footnote{When sub-sampling, we tried to preserve the proportion of the PlogP values to best get unbiased samples.}
	and we use it for the proxy evaluation of MolDQN.
	For each input, we generate 60 
	samples
	after training the model (with exploration rate set to zero) and report success if at least one of them satisfies the success criteria defined above.
	\item \textbf{VJTNN}: We sequentially optimize an input molecule using three trained models from VJTNN (models for PlogP, QED, and DRD2).
	Firstly, the PlogP model generates 20 molecules for an input molecule. We select the most similar molecules that satisfy PlogP criteria.
	Then, we repeat this process for QED and DRD2 models using the output of a preceding model as an input.
	Finally, we report success if any output of DRD2 model satisfies the success criteria.
\end{itemize}


\textbf{Result}:
We only compare VJTNN for all samples due to the infeasible running time of MolDQN as mentioned above.
As the right part of Table~\ref{tbl:single.opt} shows, the proposed model is almost two times more successful in this task.
The sub-sample experiment shows similar performance for VJTNN and ours, while MolDQN is not able to generate any successful samples.

\begin{table}[tbh]
	\centering
	\resizebox{\columnwidth}{!}{%
	\begin{tabular}{c|ccc|c|c}
		\thickhline
		\multicolumn{4}{c|}{Method}                    & Success Rate& $\pm$ \\ \hline\hline
		\multicolumn{4}{c|}{VJTNN}                     & 3.56        & -3.42      \\ \hline 
		\multirow{7}{*}{Proposed} & PNet & SNet & MBS &         &           \\\hline
		& \CheckedBox    & \CheckedBox    & \CheckedBox   & \textbf{6.98}        & -      \\
		& \CheckedBox    & \CheckedBox    & $\square$   & 6.72        & -0.26      \\
		& $\square$    & \CheckedBox    & \CheckedBox   & 6.77        & -0.21      \\
		& \CheckedBox    & $\square$     & \CheckedBox   & 6.26        & -0.72      \\
		& $\square$    & $\square$    & \CheckedBox   & 5.33        & -1.65      \\
		& $\square$    & $\square$    & $\square$   & 5.33        & -1.65     \\ \thickhline
	\end{tabular}}
	\caption{Ablation study. MBS is modified beam search, PNet is PropNet, and SNet is SimNet.}
	\label{tbl:ablation}
\end{table}


\begin{figure}[tbh]
	\centering
	\includegraphics[width=0.95\columnwidth]{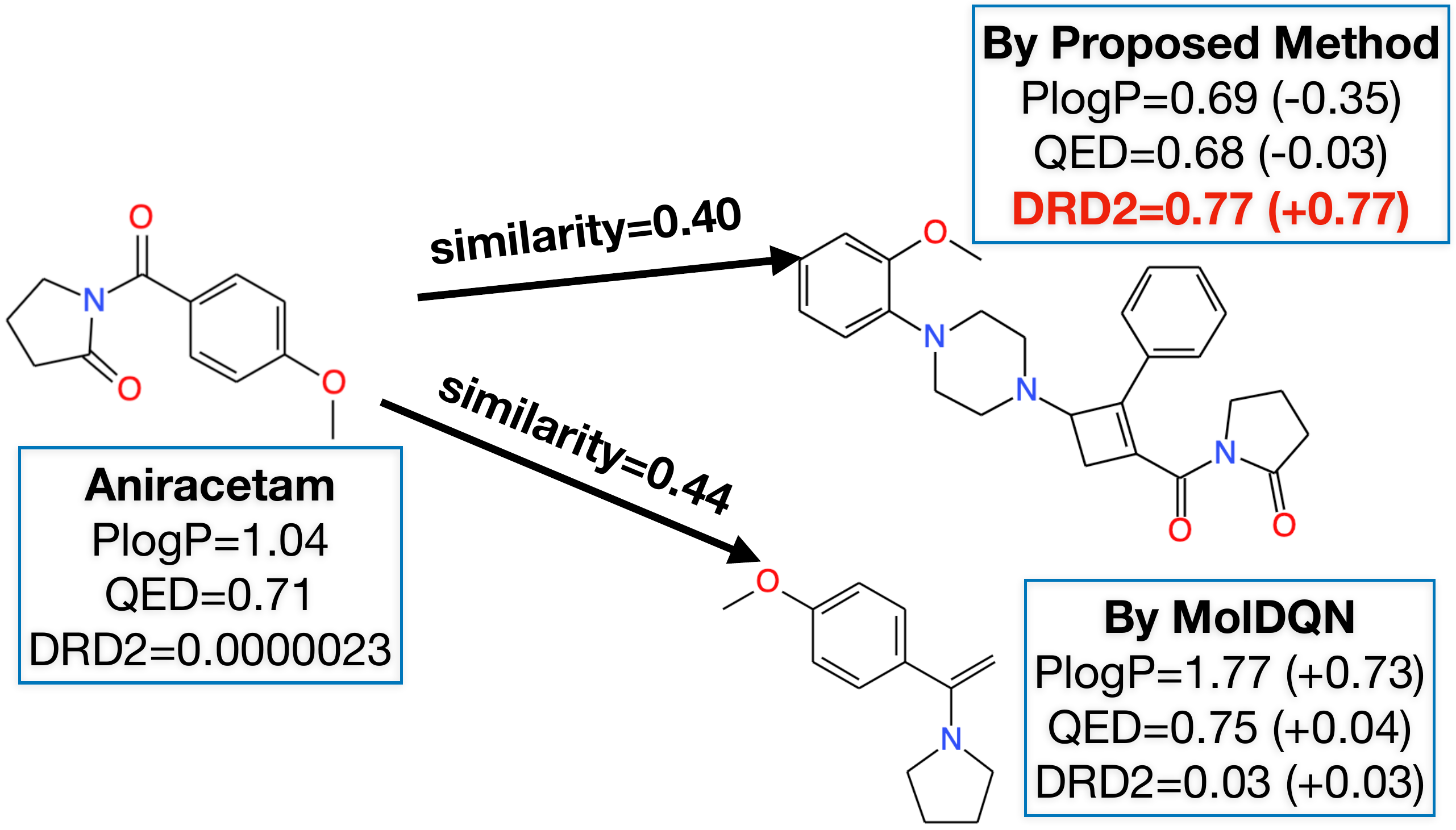}
	\caption{A case study: The molecule produced by CMG
				achieved the better DRD2 score than the molecule by MolDQN}
	\label{fig:case_study}
	 	\vspace{-1.0em}
\end{figure}

\subsection{Ablation Study}

To illustrate the effect of the two constraint networks and the modified beam search, we present the result of the ablation study in Table~\ref{tbl:ablation}.
We use the MOO task for this comparison, and the result of VJTNN is also included for the reference.
It's worthwhile to note that the proposed model without any constraint networks and the modified beam search still outperforms VJTNN by 1.77\% point. The component with the biggest contribution is SimNet that improves the performance by 0.72\% point from the model without it. 
Another interesting thing is the success rates of the last two models in Table~\ref{tbl:ablation} are identical.
The possible explanation is that if a model is trained without any constraint networks, 
the neurons generating candidate molecules could not properly convey any information about similarity and properties that can be exploited in the modified beam search.

\subsection{Case Study}
\label{ssec:case.study}


We performed a case study using an actual drug that is under the experimental stage targeting Dopamine D2 receptor (DRD2).
From Drugbank~\cite{wishart2018drugbank}, 
we first enlist all DRD2 targeting drugs that are in either experimental or investigational stages.
Among these 28 drugs, we select the lowest DRD2 scored drug, named Aniracetam (COC1=CC=C(C=C1)C(=O)N1CCCC1=O) for this study.
The goal is to improve DRD2 score with minimum perturbation of other properties.
	This can be seen as SOO, however it's a MOO, because DRD2 should be increased while others need to be unchanged.

\textbf{Baselines}:
Since one VJTNN model optimizes one property, we just run the DRD2 VJTNN model trained by Jin et al. ~\cite{jin2019learning} by feeding Aniracetam.
For MolDQN, the reward function becomes simpler as $r=\frac{1}{2}\mathbbm{1}(sim(X,Y)\ge 0.4) + \frac{1}{2}\text{DRD2}(Y)$.


\textbf{Result}:
In Figure~\ref{fig:case_study}, we compare the molecules generated by MolDQN (C=C(c1ccc(OC) cc1)N1CCCC1) 
and ours (COc1ccccc1N1CCN(C2CC (C(=O)N3CCCC3=O)=C2c2ccccc2)CC1), 
excluding the result of VJTNN,
because VJTNN didn't generate valid ($sim(X,Y) \ge 0.4$) molecules.
In terms of the predicted DRD2 scores, our molecule reached 0.77 whereas MolDQN's molecule only recorded 0.03.
For the other two properties which should be unchanged, our molecule seems to be stable with changes in PlogP by -0.35 and QED by -0.03 when compared with the MolDQN molecule that showed larger changes especially in PlogP.
Although one case study cannot prove the general superiority of the proposed model, the proposed model consistently outperforms other baselines in all benchmarks (SOO, MOO, and the case study).

\section{Conclusion}
\label{sec:conclusion}
This paper proposes a new controlled molecule generation model using the self-attention  based molecule translation model and two constraint networks. We pre-train and transfer the weights of the two constraint networks so that they can effectively regulate the output molecules. Not only that, we present a new beam search algorithm using these networks.
Experimental results show that the proposed model outperforms all other baseline approaches in both single-objective optimization and multi-objective optimization by a large margin.
Moreover, the case study using an actual experimental drug shows the practicality of the proposed model.
In the ablation study, we present how each sub-unit contributes to model performance.
It's worth to note that our model is trained once and evaluated for all tasks (SOO, MOO and the case study), which shows practicality and generalizability.

\bibliographystyle{siamplain}
\bibliography{sdmcmg.bib}

\begin{thebibliography}{10}

\bibitem{bahdanau2014neural}
{\sc D.~Bahdanau, K.~Cho, and Y.~Bengio}, {\em Neural machine translation by
  jointly learning to align and translate}, arXiv preprint arXiv:1409.0473,
  (2014).

\bibitem{bickerton2012quantifying}
{\sc G.~R. Bickerton, G.~V. Paolini, J.~Besnard, S.~Muresan, and A.~L.
  Hopkins}, {\em Quantifying the chemical beauty of drugs}, Nature chemistry, 4
  (2012), p.~90.

\bibitem{bjerrum2017molecular}
{\sc E.~J. Bjerrum and R.~Threlfall}, {\em Molecular generation with recurrent
  neural networks (rnns)}, arXiv preprint arXiv:1705.04612,  (2017).

\bibitem{dai2016discriminative}
{\sc H.~Dai, B.~Dai, and L.~Song}, {\em Discriminative embeddings of latent
  variable models for structured data}, in International conference on machine
  learning, 2016, pp.~2702--2711.

\bibitem{dai2018syntax}
{\sc H.~Dai, Y.~Tian, B.~Dai, S.~Skiena, and L.~Song}, {\em Syntax-directed
  variational autoencoder for structured data}, arXiv preprint
  arXiv:1802.08786,  (2018).

\bibitem{dalke2018mmpdb}
{\sc A.~Dalke, J.~Hert, and C.~Kramer}, {\em mmpdb: An open-source matched
  molecular pair platform for large multiproperty data sets}, Journal of
  chemical information and modeling, 58 (2018), pp.~902--910.

\bibitem{devlin2018bert}
{\sc J.~Devlin, M.-W. Chang, K.~Lee, and K.~Toutanova}, {\em Bert: Pre-training
  of deep bidirectional transformers for language understanding}, arXiv
  preprint arXiv:1810.04805,  (2018).

\bibitem{dimasi2016innovation}
{\sc J.~A. DiMasi, H.~G. Grabowski, and R.~W. Hansen}, {\em Innovation in the
  pharmaceutical industry: new estimates of r\&d costs}, Journal of health
  economics, 47 (2016), pp.~20--33.

\bibitem{dossetter2013matched}
{\sc A.~G. Dossetter, E.~J. Griffen, and A.~G. Leach}, {\em Matched molecular
  pair analysis in drug discovery}, Drug Discovery Today, 18 (2013),
  pp.~724--731.

\bibitem{ertl2017silico}
{\sc P.~Ertl, R.~Lewis, E.~Martin, and V.~Polyakov}, {\em In silico generation
  of novel, drug-like chemical matter using the lstm neural network}, arXiv
  preprint arXiv:1712.07449,  (2017).

\bibitem{ghifary2016deep}
{\sc M.~Ghifary, W.~B. Kleijn, M.~Zhang, D.~Balduzzi, and W.~Li}, {\em Deep
  reconstruction-classification networks for unsupervised domain adaptation},
  in European Conference on Computer Vision, Springer, 2016, pp.~597--613.

\bibitem{gomez2018automatic}
{\sc R.~G{\'o}mez-Bombarelli, J.~N. Wei, D.~Duvenaud, J.~M.
  Hern{\'a}ndez-Lobato, B.~S{\'a}nchez-Lengeling, D.~Sheberla,
  J.~Aguilera-Iparraguirre, T.~D. Hirzel, R.~P. Adams, and A.~Aspuru-Guzik},
  {\em Automatic chemical design using a data-driven continuous representation
  of molecules}, ACS central science, 4 (2018), pp.~268--276.

\bibitem{griffen2011matched}
{\sc E.~Griffen, A.~G. Leach, G.~R. Robb, and D.~J. Warner}, {\em Matched
  molecular pairs as a medicinal chemistry tool: miniperspective}, Journal of
  medicinal chemistry, 54 (2011), pp.~7739--7750.

\bibitem{guimaraes2017objective}
{\sc G.~L. Guimaraes, B.~Sanchez-Lengeling, C.~Outeiral, P.~L.~C. Farias, and
  A.~Aspuru-Guzik}, {\em Objective-reinforced generative adversarial networks
  (organ) for sequence generation models}, arXiv preprint arXiv:1705.10843,
  (2017).

\bibitem{hochreiter1997long}
{\sc S.~Hochreiter and J.~Schmidhuber}, {\em Long short-term memory}, Neural
  computation, 9 (1997), pp.~1735--1780.

\bibitem{howard2018universal}
{\sc J.~Howard and S.~Ruder}, {\em Universal language model fine-tuning for
  text classification}, arXiv preprint arXiv:1801.06146,  (2018).

\bibitem{hu2017toward}
{\sc Z.~Hu, Z.~Yang, X.~Liang, R.~Salakhutdinov, and E.~P. Xing}, {\em Toward
  controlled generation of text}, in Proceedings of the 34th International
  Conference on Machine Learning-Volume 70, JMLR. org, 2017, pp.~1587--1596.

\bibitem{jaitly2012application}
{\sc N.~Jaitly, P.~Nguyen, A.~Senior, and V.~Vanhoucke}, {\em Application of
  pretrained deep neural networks to large vocabulary speech recognition},
  (2012).

\bibitem{jin2018junction}
{\sc W.~Jin, R.~Barzilay, and T.~Jaakkola}, {\em Junction tree variational
  autoencoder for molecular graph generation}, arXiv preprint arXiv:1802.04364,
   (2018).

\bibitem{jin2017predicting}
{\sc W.~Jin, C.~Coley, R.~Barzilay, and T.~Jaakkola}, {\em Predicting organic
  reaction outcomes with weisfeiler-lehman network}, in Advances in Neural
  Information Processing Systems, 2017, pp.~2607--2616.

\bibitem{jin2019learning}
{\sc W.~Jin, K.~Yang, R.~Barzilay, and T.~Jaakkola}, {\em Learning multimodal
  graph-to-graph translation for molecular optimization}, ICLR,  (2019).

\bibitem{kirkpatrick2004chemical}
{\sc P.~Kirkpatrick and C.~Ellis}, {\em Chemical space}, 2004.

\bibitem{kusner2017grammar}
{\sc M.~J. Kusner, B.~Paige, and J.~M. Hern{\'a}ndez-Lobato}, {\em Grammar
  variational autoencoder}, in Proceedings of the 34th International Conference
  on Machine Learning-Volume 70, JMLR. org, 2017, pp.~1945--1954.

\bibitem{li2018multi}
{\sc Y.~Li, L.~Zhang, and Z.~Liu}, {\em Multi-objective de novo drug design
  with conditional graph generative model}, Journal of cheminformatics, 10
  (2018), p.~33.

\bibitem{lu2013speech}
{\sc X.~Lu, Y.~Tsao, S.~Matsuda, and C.~Hori}, {\em Speech enhancement based on
  deep denoising autoencoder.}, in Interspeech, 2013, pp.~436--440.

\bibitem{medress1977speech}
{\sc M.~F. Medress, F.~S. Cooper, J.~W. Forgie, C.~Green, D.~H. Klatt, M.~H.
  O'Malley, E.~P. Neuburg, A.~Newell, D.~Reddy, B.~Ritea, et~al.}, {\em Speech
  understanding systems: Report of a steering committee}, Artificial
  Intelligence, 9 (1977), pp.~307--316.

\bibitem{olivecrona2017molecular}
{\sc M.~Olivecrona, T.~Blaschke, O.~Engkvist, and H.~Chen}, {\em Molecular
  de-novo design through deep reinforcement learning}, Journal of
  cheminformatics, 9 (2017), p.~48.

\bibitem{polishchuk2013estimation}
{\sc P.~G. Polishchuk, T.~I. Madzhidov, and A.~Varnek}, {\em Estimation of the
  size of drug-like chemical space based on gdb-17 data}, Journal of
  computer-aided molecular design, 27 (2013), pp.~675--679.

\bibitem{rogers2010extended}
{\sc D.~Rogers and M.~Hahn}, {\em Extended-connectivity fingerprints}, Journal
  of chemical information and modeling, 50 (2010), pp.~742--754.

\bibitem{rothe2015dex}
{\sc R.~Rothe, R.~Timofte, and L.~Van~Gool}, {\em Dex: Deep expectation of
  apparent age from a single image}, in Proceedings of the IEEE international
  conference on computer vision workshops, 2015, pp.~10--15.

\bibitem{samanta2018designing}
{\sc B.~Samanta, A.~De, N.~Ganguly, and M.~Gomez-Rodriguez}, {\em Designing
  random graph models using variational autoencoders with applications to
  chemical design}, arXiv preprint arXiv:1802.05283,  (2018).

\bibitem{sanchez2017optimizing}
{\sc B.~Sanchez-Lengeling, C.~Outeiral, G.~L. Guimaraes, and A.~Aspuru-Guzik},
  {\em Optimizing distributions over molecular space. an objective-reinforced
  generative adversarial network for inverse-design chemistry (organic)},
  (2017).

\bibitem{segler2018generating}
{\sc M.~H. Segler, T.~Kogej, C.~Tyrchan, and M.~P. Waller}, {\em Generating
  focused molecule libraries for drug discovery with recurrent neural
  networks}, ACS central science, 4 (2018), pp.~120--131.

\bibitem{shanmugasundaram2016monitoring}
{\sc V.~Shanmugasundaram, L.~Zhang, S.~Kayastha, A.~de~la Vega~de Leon,
  D.~Dimova, and J.~Bajorath}, {\em Monitoring the progression of
  structure--activity relationship information during lead optimization},
  Journal of medicinal chemistry, 59 (2016), pp.~4235--4244.

\bibitem{shin2017classification}
{\sc B.~Shin, F.~H. Chokshi, T.~Lee, and J.~D. Choi}, {\em Classification of
  radiology reports using neural attention models}, in 2017 International Joint
  Conference on Neural Networks (IJCNN), IEEE, 2017, pp.~4363--4370.

\bibitem{pmlr-v106-shin19a}
{\sc B.~Shin, S.~Park, K.~Kang, and J.~C. Ho}, {\em Self-attention based
  molecule representation for predicting drug-target interaction}, in
  Proceedings of the 4th Machine Learning for Healthcare Conference, vol.~106
  of Proceedings of Machine Learning Research, PMLR, 09--10 Aug 2019,
  pp.~230--248.

\bibitem{sterling2015zinc}
{\sc T.~Sterling and J.~J. Irwin}, {\em Zinc 15--ligand discovery for
  everyone}, Journal of chemical information and modeling, 55 (2015),
  pp.~2324--2337.

\bibitem{vaswani2017attention}
{\sc A.~Vaswani, N.~Shazeer, N.~Parmar, J.~Uszkoreit, L.~Jones, A.~N. Gomez,
  {\L}.~Kaiser, and I.~Polosukhin}, {\em Attention is all you need}, in
  Advances in neural information processing systems, 2017, pp.~5998--6008.

\bibitem{vogt2018computational}
{\sc M.~Vogt, D.~Yonchev, and J.~Bajorath}, {\em Computational method to
  evaluate progress in lead optimization}, Journal of medicinal chemistry, 61
  (2018), pp.~10895--10900.

\bibitem{weininger1988smiles}
{\sc D.~Weininger}, {\em Smiles, a chemical language and information system. 1.
  introduction to methodology and encoding rules}, Journal of chemical
  information and computer sciences, 28 (1988), pp.~31--36.

\bibitem{wishart2018drugbank}
{\sc D.~S. Wishart, Y.~D. Feunang, A.~C. Guo, E.~J. Lo, A.~Marcu, J.~R. Grant,
  T.~Sajed, D.~Johnson, C.~Li, Z.~Sayeeda, et~al.}, {\em Drugbank 5.0: a major
  update to the drugbank database for 2018}, Nucleic acids research, 46 (2018),
  pp.~D1074--D1082.

\bibitem{yang2017chemts}
{\sc X.~Yang, J.~Zhang, K.~Yoshizoe, K.~Terayama, and K.~Tsuda}, {\em Chemts:
  an efficient python library for de novo molecular generation}, Science and
  technology of advanced materials, 18 (2017), pp.~972--976.

\bibitem{you2018graph}
{\sc J.~You, B.~Liu, Z.~Ying, V.~Pande, and J.~Leskovec}, {\em Graph
  convolutional policy network for goal-directed molecular graph generation},
  in Advances in neural information processing systems.

\bibitem{zhou2019optimization}
{\sc Z.~Zhou, S.~Kearnes, L.~Li, R.~N. Zare, and P.~Riley}, {\em Optimization
  of molecules via deep reinforcement learning}, Scientific reports, 9 (2019),
  pp.~1--10.

\end{thebibliography}

\end{document}